\documentclass{article}
\usepackage{ijcai24}

\usepackage{times}
\usepackage{soul}
\usepackage{url}
\usepackage[hidelinks]{hyperref}
\usepackage[utf8]{inputenc}
\usepackage[small]{caption}
\usepackage{graphicx}
\usepackage{amsmath}
\usepackage{amsthm}
\usepackage{booktabs}
\usepackage{algorithm}
\usepackage{algorithmic}
\usepackage[switch]{lineno}

\usepackage{todonotes}
\usepackage{subcaption}
\usepackage{natbib}

\usepackage{stfloats}

\title{Case-Enhanced Vision Transformer: \\ Improving Explanations of Image Similarity with a ViT-based Similarity Metric}

\author{
Ziwei Zhao$^1$
\and
David Leake$^1$\and
Xiaomeng Ye$^2$\And
David Crandall$^1$\\
\affiliations
$^1$Indiana University, Bloomington IN 47408, USA\\
$^2$Berry College, Mount Berry GA 30149, USA\\
\emails
\{zz47, leake, djcran\}@iu.edu,
xye@berry.edu
}

\begin{document}

\maketitle

\begin{abstract}
This short paper presents preliminary research on the Case-Enhanced Vision Transformer (CEViT), a similarity measurement method aimed at improving the explainability of similarity assessments for image data. Initial experimental results suggest that integrating CEViT into k-Nearest Neighbor (k-NN) classification yields classification accuracy comparable to state-of-the-art computer vision models, while adding capabilities for illustrating differences between classes. CEViT explanations can be influenced by prior cases, to illustrate aspects of similarity relevant to those cases. \footnote{This paper has been accepted at the IJCAI 2024
Workshop on Explainable Artificial Intelligence (XAI).}

\end{abstract}

\section{Introduction}

In computer vision, Convolutional Neural Networks (CNNs) for image classification have achieved remarkable performance, but are black-box systems that cannot be explained directly \citep{adadi2018peeking}. In contrast, traditional machine learning algorithms such as k-Nearest Neighbors (k-NN) are inherently interpretable, due to the ability to present the examples on which decisions are based. However, their accuracy tends to lag behind that of modern ML models.

The interpretability of k-NN stems from its reliance on existing data. This process is analogous to Case-Based Reasoning (CBR), a traditional AI paradigm that involves solving new problems by recalling and adapting solutions from past experiences, or "cases". The interpretability of CBR systems has long been seen as an important strength, and human subjects studies support that cases provide compelling explanations.
\citep{cunningham-doyle-loughrey03,gates2023cases}. 

The core factor impacting the performance of k-NN is the similarity measurement function it uses. 
Traditional methods for determining similarity, such as L1 or L2 distance, prove inadequate for image data. This is because two images may not align perfectly, and attempting to explain visually by contrasting two images in pixel space results in a blurry image lacking explanatory information. Similarity measurement is also important for CBR studies to  explain why a particular case was retrieved \citep{massie-craw-wiratunga04}, and for cases with feature-vector problem representations, explanations may be based on feature-by-feature comparisons. 
This approach is not useful for images processed on the pixel level.

\begin{figure}
     \centering
     \begin{subfigure}[b]{0.235\textwidth}
         \centering
         \includegraphics[width=\textwidth]{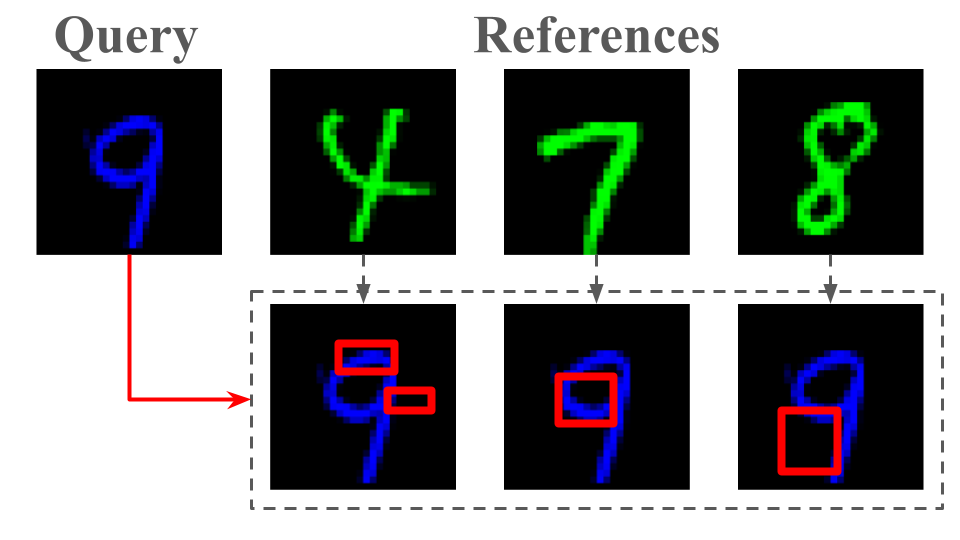}
         \caption{\textbf{Human explanations.}}
         \label{fig:overview_human}
     \end{subfigure}
     \hfill
     \begin{subfigure}[b]{0.235\textwidth}
         \centering
         \includegraphics[width=\textwidth]{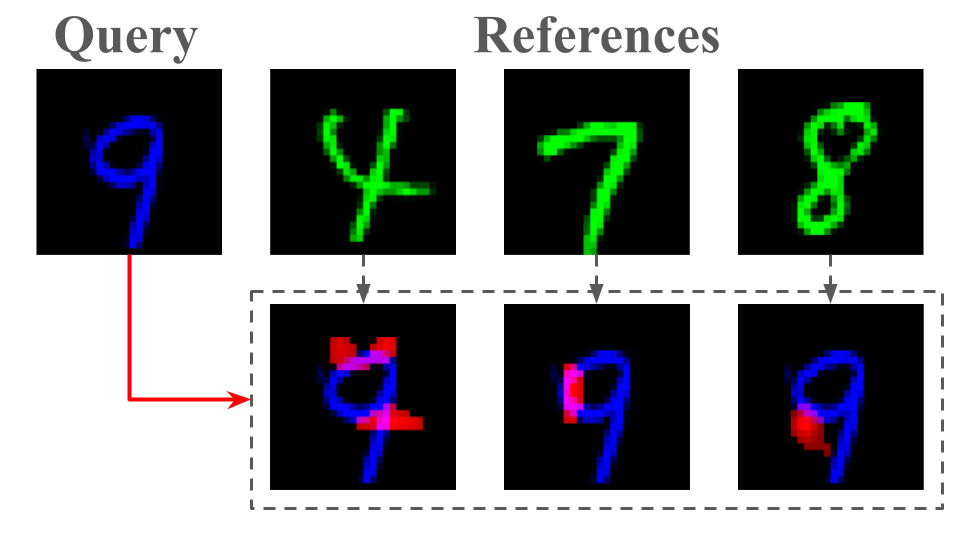}
         \caption{\textbf{CEViT explanations.}}
         \label{fig:overview_cevit}
     \end{subfigure}
    \caption{\textbf{Explaining class differences.}}
    \label{fig:overview}
    \vspace{-15pt}
\end{figure}

Humans excel at explaining the differences between images concisely and accurately, commonly using more abstract image characteristics to underpin their explanations.   For instance, 
as depicted in Figure \ref{fig:overview_human}, it is plausible that
the difference between an image of the digit 7 and an image of the digit 9 as characterized by a human would lie in the upper stroke, as shown by the red boxes. 
Recent advances in computer vision include highly accurate similarity measurement models, such as Siamese Neural Networks \citep{koch2015siamese,LealTaix2016LearningBT}. Despite their high performance, these methods share the same issue that they operate as black boxes. They commonly use convolutional Neural Networks (CNNs) to generate features for determining similarity resulting in two potential pitfalls:  the process of feature generation lacks explainability, and the features themselves may have limited visual connection to the original image.

The Vision Transformer (ViT) \citep{dosovitskiy2020image} stands out as a revolutionary architecture. Unlike conventional CNNs, ViT uses a transformer architecture \citep{vaswani2017attention}, enabling it to capture dependencies across subpatches of an image. This approach has yielded remarkable performance across various computer vision tasks, such as classification, object detection, and segmentation.

The core of a transformer model is the Query-Key-Value (QKV) attention mechanism. The QKV mechanism computes attention scores between query (Q) and keys (K), determining the relevance and importance of each value (V) with respect to the query. The attention scores (Q$\times$K) offer great potential for explainability, as they naturally highlight the parts of the query that are important during the process. Unfortunately, in traditional classification tasks, these attention scores may have limited explanatory value. This limitation arises from the system's need to consider every possible class during decision-making, often resulting in a widespread attention distribution over the entire input.

Drawing inspiration from CBR, this paper presents the Case-Enhanced Vision Transformer (CEViT), a ViT-based similarity measurement method aimed at enhancing similarity assessment for image data. 
Initial experimental results suggest that a k-NN model using CEViT as the similarity metric offers two potential explainability advantages compared to image classification using CNNs and ViT:
\begin{itemize}
\item It yields comparable classification accuracy to state-of-the-art computer vision methods, 
while retaining the inherent benefit of utilizing reference cases.
\item The attention scores generated by CEViT can effectively illustrate the differences between classes, as depicted in Figure \ref{fig:overview_cevit}. 
\end{itemize}
The code for CEViT is publicly available at \url{https://github.com/ziweizhao1993/CEViT}.

\section{Related Work}

\subsection{Explainable AI for Computer Vision}

Modern computer vision techniques have gained huge success through the use of deep learning models. 
These models often operate as black boxes due to their vast number of parameters and complex layer connections, making it challenging to understand their inner function. 
Tremendous effort has been devoted to opening the black box by explaining their decision making process. 
As some examples, Saliency maps highlight regions in an image that are most activated for the system's prediction \citep{Simonyan2013DeepIC}; Grad-CAM uses the gradients of a target concept (e.g., a dog in an image) in convolutional layers to produce a coarse localization map \citep{8237336}; Attention mechanisms assign weights to parts of the input image and higher weights indicate more importance in the model's decision \citep{Guo2022}.

\subsection{Applying CBR to Image Data}

CBR can be applied to many tasks that involve image inputs. In contrast to common computer vision techniques, CBR methods can provide similar prior cases as explanations of current decisions. \cite{ye2021applying} applied the Case Difference Heuristic (CDH) approach for learning case adaptations to predict age from facial images. The Class-to-Class Variational Autoencoder~\citep{ye2022generation,zhao2022generating} learns inter-class patterns, showing potential in few-shot image generation and  counterfactual visual explanations. \cite{kenny2020generating} proposed PIECE, a method that modifies ``exceptional" features in images to generate counterfactual explanations for black-box CNN classifiers. Kenny and collaborators \citep{kenny2019twin,kenny2021explaining} introduced a twin-system that pairs a black box Neural Network with a white box CBR system to identify nearest neighbor cases explaining image classification outcomes.  \cite{ye2020applying} integrates CBR with a Siamese Network for image classification and discovering prototypical cases.
Our work contrasts with previous research by using the Vision Transformer, a novel backbone model in computer vision, instead of a traditional CNN.

\subsection{Similarity Metrics in CBR}

Case retrieval in CBR relies on a similarity metric to determine the distances between cases. For simple data sets in which the problem addressed by the case is described as a feature vector, Euclidean distance may suffice. Metric learning methods such as NCA \citep{NIPS2004_42fe8808} and LMNN \citep{10.5555/1577069.1577078} transforming the input space to enhance accuracy. \citet{Mathisen19} proposed and extended siamese neural network that learns an embedding of cases and calculates distances between cases. For high dimensional, complex, or structural data (e.g. audio, video, image, text), deep learning models are often used to extract features to which the similarity metric is applied \citep{Sani2017,Wilkerson2021OnCK,Turner2018}. 

\section{The CEViT Method}
In this section, we begin by explaining the design of our transformer model (CEViT). Then, we show the procedure for accessing the attention mask of the transformer model to enhance model explainability. Finally, we present a classification pipeline that combines CEViT with k-NN by using CEViT as the distance metric for k-NN.

\subsection{Model design}
\begin{figure}
    \centering
    \includegraphics[width=0.48\textwidth]{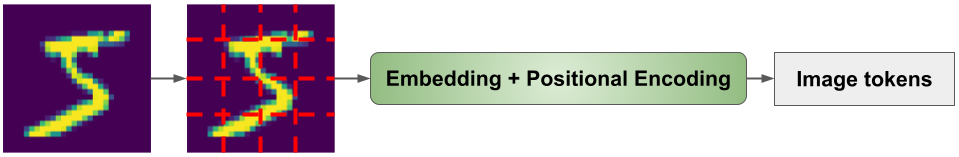}
    \caption{\textbf{The patchify process}}
    \label{fig:patchify}
    \vspace{-15pt}
\end{figure}

We follow the model design of the Vision Transformer (ViT) introduced in \cite{dosovitskiy2020image}. Traditionally, ViT comprises 3 modules:

\begin{itemize}
    \item \textbf{Patchify process:} It splits the input image into smaller patches and processes them into tokens via linear projection and positional embedding, as shown in Figure \ref{fig:patchify}.
    \item \textbf{Transformer module:} This module comprises encoder blocks consisting of normalization, MLP (Multi-Layer Perceptron), and multi-head attention layers.
    \item \textbf{MLP head:} This produces the desired output, such as class labels ranging from 0 to 9 for digit recognition task.
\end{itemize}

\begin{figure}
     \centering
     \hfill
     \begin{subfigure}[b]{0.1535\textwidth}
         \centering
         \includegraphics[width=\textwidth]{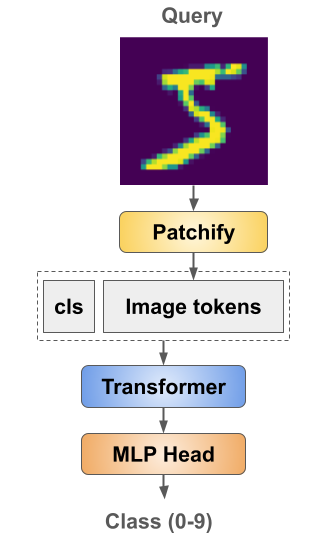}
         \caption{\textbf{Traditional ViT}}
         \label{fig:vit}
     \end{subfigure}
     \hfill
     \begin{subfigure}[b]{0.28\textwidth}
         \centering
         \includegraphics[width=\textwidth]{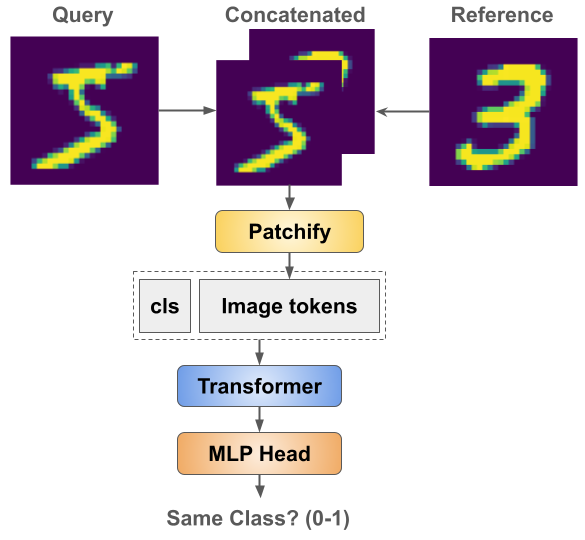}
         \caption{\textbf{Case-Enhanced ViT (CEViT)}}
         \label{fig:five over x}
     \end{subfigure}
    \caption{Given an input image (or image-like tensor), the patchify process divides it into $N^2$ smaller image patches (in this example, $N=4$). These image tokens, along with a classification token, are then fed into the transformer model to produce the output.}
    \label{fig:three graphs}
\end{figure}

Our method, CEViT, modifies ViT in two ways. First, instead of taking a single h$\times$w$\times$c image as input, CEViT concatenates the input image with a reference image along the channel dimension, resulting in an image-like tensor with dimensions h$\times$w$\times$2c. Second, we modify the MLP head to produce a single score between 0 and 1, indicating the likelihood that the two images belong to the same class. Figure \ref{fig:three graphs} illustrates the ViT and CEViT processes. For an explanation of each ViT module, please see \cite{dosovitskiy2020image}.

\subsection{Attention Mask}
\begin{figure}[t!]
\vspace{-10pt}
\centering
\includegraphics[width=0.4\textwidth]{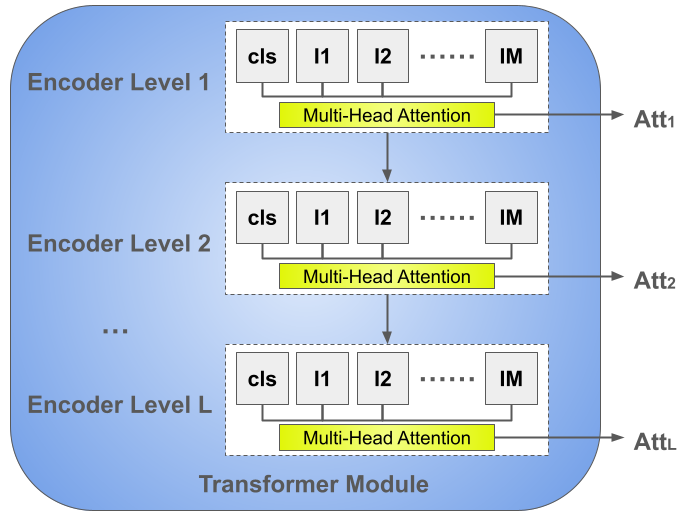}
\caption{Accessing the attention mask between the classification token and M image tokens in a transformer model with L encoders. }
\label{fig:attention}
\vspace{-15pt}
\end{figure}

During the inference of a ViT model, each encoder block computes an attention mask between tokens. Typically, ViT includes an additional CLS (classification) token, utilizing it as an aggregate representation for all tokens to generate the final prediction. Consequently, we can trace back the attention masks at each encoder level to observe how the CLS token attended to the image tokens, with each token representing a region in the original image, as shown in Figure \ref{fig:attention}. Hence, these attention masks can be used to visualize how each region of the original image influenced the final classification.

In CEViT, by augmenting the input image with the reference image, the attention masks visualize the regions that influenced the similarity or dissimilarity between the query image and the reference image. This visual representation is particularly valuable for explaining classification differences.

\subsection{Using CEViT for classification}

To employ CEViT for classification, we integrate it into a k-NN 
system by running CEViT between the query image and reference images from each class. The K samples with the highest similarity scores then vote for the final prediction.

\section{Experiments}
We evaluated the classification accuracy and explainability of CEViT against a standard ViT and k-NN on the MNIST \cite{lecun1998gradient} dataset.

\subsection{Implementation Details}
We implemented both ViT and CEViT using the Pytorch framework \citep{Paszke_PyTorch_An_Imperative_2019} and the Torchvision  library \citep{TorchVision_maintainers_and_contributors_TorchVision_PyTorch_s_Computer_2016}. To ensure a fair comparison, we used the same backbone for both ViT and CEViT, incorporating a patchify module that subdivides the input into 49 ($7\times7$) small patches, and a transformer module consisting of 6 encoder layers.

During training, CEViT was given pairs of images, each with a 50\% probability of being from the same class. The output of CEViT is a score between 0 and 1, indicating the likelihood that the images belong to the same class. We then applied cross-entropy loss, where a ground truth label of 0 (resp.\ 1) indicates different (resp.\ same) classes.

Both ViT and CEViT were trained for 200 epochs using the AdamW optimizer \citep{loshchilov2017decoupled} with an empirically set learning rate of 0.001. We decreased the rate by 10\% per 20 epochs. For classification, we integrated CEViT into a k-NN with K=15, chosen empirically. 

\subsection{Quantitative Evaluation}
We first compare the fundamental performance of ViT, k-NN with Euclidean distance, and k-NN using CEViT as similarity measure. 
As shown in Table \ref{tab:1}, the classification accuracy of CEViT+k-NN is higher than k-NN and comparable to that of ViT. Meanwhile, CEViT preserves the advantage of k-NN by providing nearest cases as explanations and can generate attention masks more related to class differences, potentially leading to better explainability.

\begin{table*}
    \centering
    \begin{tabular}{|c|c|c|c|c|}
        \hline
        Method & Classification accuracy & Nearest neighbors & Attention mask & Inter-class attention mask\\
        \hline
        k-NN & 97.1\% & \textbf{Yes} & No & No\\
        \hline
        ViT & \textbf{99.1\%} & No & \textbf{Yes} & No\\
        \hline
        CEViT+k-NN & \textbf{99.0\%} & \textbf{Yes} & \textbf{Yes} & \textbf{Yes}\\
        \hline
    \end{tabular}
    \caption{Comparison between k-NN, ViT and CEViT. CEViT achieved similar accuracy on MNIST compared to ViT, while also having better explainability potential.}
    \label{tab:1}
\end{table*}

We also introduce a quantitative evaluation method to assess the explainability of the generated attention masks. The method is based on the premise that the parts of an image contributing to class differences best explain classifications, and that the extent to which a mask identifies such parts can be judged by its ability to guide the modification of a query image to match a distractor class.

\begin{figure}[h!]

\centering
\vspace{-1pt}
\includegraphics[width=0.48\textwidth]{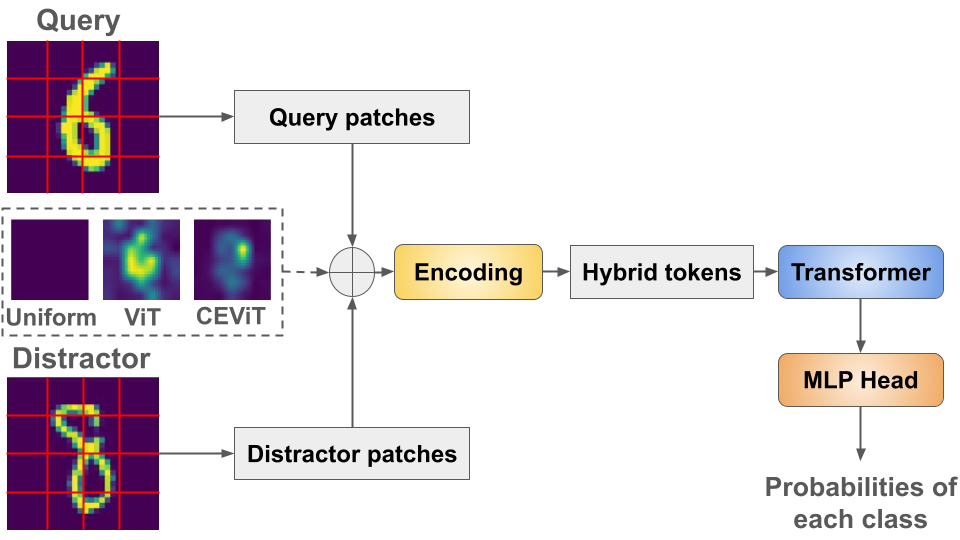}
\caption{\textbf{The quantitative evaluation process.} 
Following the patchify process, image patches from both the query image and the distractor image are merged based on the normalized masks. These hybrid tokens are used to compute updated class likelihoods.}
\label{fig:quantitative}
\vspace{-5pt}
\end{figure}

\begin{table*}
    \centering
    \begin{tabular}{|c|c|c|c|c|c|c|}
        \hline
        Method & ViT(first layer) & ViT(last layer) & ViT(Avg.) & CEViT(first layer) & CEViT(last layer) & CEViT(Avg.)\\
        \hline
        Distraction score$\uparrow$& -0.0030 & -0.0062 & -0.0037 & 0.0286 & 0.0368 & 0.0259\\
        \hline
    \end{tabular}
    \caption{\textbf{Results of our quantitative evaluation.} 
Results from our quantitative evaluation indicate that ViT performs comparably to the baseline (a uniform mask). However, the masks generated by CEViT are more effective in diverting the query image towards the distractor class. This suggests that CEViT is better at illustrating the differences between classes.}
    \label{tab:2}
    
\end{table*}

As depicted in Figure \ref{fig:quantitative}, we start by selecting a query image with class $q$ and a distractor image with class $d$. Subsequently, we conduct classification of the query image using both ViT and CEViT, resulting in attention masks $M_{V}$ and $M_{C}$. Additionally, we employ a uniform mask $M_{U}$ with consistent values across all pixels. The uniform mask $M_{U}$ serves as a baseline, against which we assess the relative performance improvements of ViT and CEViT. All three attention masks are normalized to have the same mean value $\mu$. After the patchify process which divides the images into $N^2$ smaller image patches, we acquire image patches $P_{q1} ... P_{qN^2}$ and $P_{d1} ... P_{dN^2}$, and merge them into hybrid patches $P_{h1} ... P_{hN^2}$ using the following equation:

\begin{equation}
P_{hi} = (1-M_{i})P_{qi} + M_{i}P_{di} \qquad i\in{\{1, ... ,N^2\}}
\end{equation}

In the equation above, $M_{i}$ represents the value in mask $M$ corresponding to the $i$th image patch, indicating the importance of that patch toward the final prediction. 

After the merge step, we use them to generate hybrid image tokens and feed the three sets of hybrid tokens created by $M_{V}$, $M_{C}$, and $M_{U}$ into the transformer module to compute the likelihoods for each class, resulting in $P_{V}$, $P_{U}$, and $P_{C}$ respectively. Subsequently, we use the following equation to determine the distraction scores for ViT and CEViT:
\vspace{-2pt}
\begin{align}
S_{V} = P_{V}(d) - P_{U}(d)\\
S_{C} = P_{C}(d) - P_{U}(d)
\label{eq:1}
\end{align}

To better explain this process, consider the example depicted in Figure \ref{fig:quantitative}. Here, we have a query image with class $q=6$ and a distractor image with class $d=8$. Upon merging their image patches, guided by the masks, we recompute the probability of each class. By ensuring the query image is modified by an equal degree (achieved through normalization of all masks to maintain the same mean value $\mu$), the hybrid tokens generated from masks that effectively illustrate the difference between classes 6 and 8 should exhibit a higher probability for class 8 (the distractor class). 

We tested six attention masks generated by ViT and CEViT from the first encoder layer, last encoder layer, and the average of all layers on 10 hand-selected pairs of similar MNIST classes (e.g., 1 and 7), with scores averaged. 
Table \ref{tab:2} shows that CEViT's attention masks are better than ViT's in modifying images towards the distractor class, demonstrating our model's enhanced capability for explaining class differences.

\begin{figure}
\vspace{-5pt}
    \centering
    \includegraphics[width=0.4\textwidth]{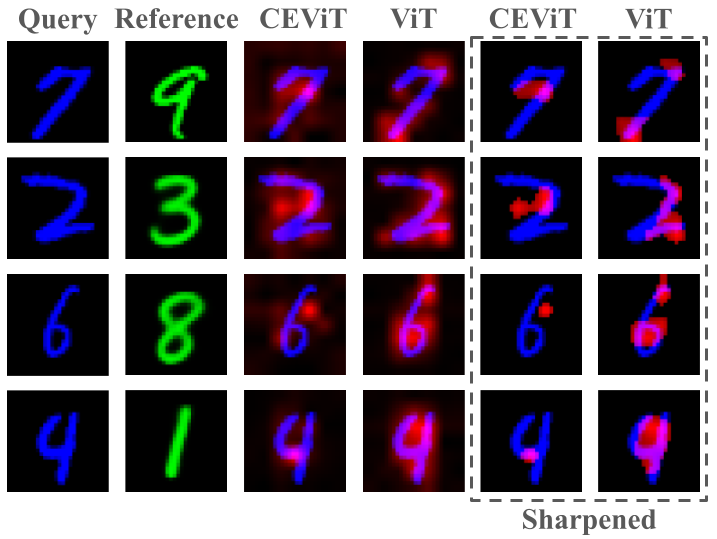}
    \caption{\textbf{Qualitative Results}}
    \label{fig:newqualitative}
\vspace{-15pt}
\end{figure}

\subsection{Qualitative Evaluation}
\label{section 4.3}
We present several qualitative results in Figure \ref{fig:newqualitative}. To enhance visualization, we sharpened the attention masks of both CEViT and ViT by scaling them to a range of 0-1 and subsequently filtering out pixels with values less than 0.5. As shown in Figure \ref{fig:newqualitative}, the attention masks generated by CEViT are notably more focused and better illustrate class differences compared to those generated by ViT. The supplementary materials provide additional qualitative samples.

\section{Conclusion and Future Work}
In this paper, we introduce the Case-Enhanced Vision Transformer (CEViT), a novel similarity metric for the CBR retrieval process. Initial experiments show encouraging classification performance for CEViT combined with KNN, along with the capability to explain class differences by generating attention masks. We highlight two directions for future research: Investigating CEViT's performance on more challenging dataset, and using CEViT's capabilities to assist in generating counterfactual and semi-factual explanations. 

\section{Acknowledgements}
We thank Lawrence Gates for his helpful comments on a draft of this paper. 

\bibliographystyle{named}
\bibliography{ijcai24,bibliography,ai,ai2}

\clearpage
\section*{Supplementary Material} 

For our quantitative evaluation, we selected 10 pairs of classes spanning every class of the MNIST dataset, with each pair representing two classes with notable similarity (e.g., 1 and 7), as shown below:

\begin{itemize}
    \item \textbf{Queries}: 0,1,2,3,4,5,6,7,8,9
    \item \textbf{Distractors}: 6,7,3,5,1,8,0,2,9,4
\end{itemize}

We did this because we expect it to be more difficult to explain the difference between two classes when they are similar. For example, it is more challenging to explain the difference between a white Persian cat and a white Ragdoll cat than to explain the difference between a white Persian cat and a black dog. 

\begin{figure*}[!b]
    \centering
    \includegraphics[width=0.4\textwidth]{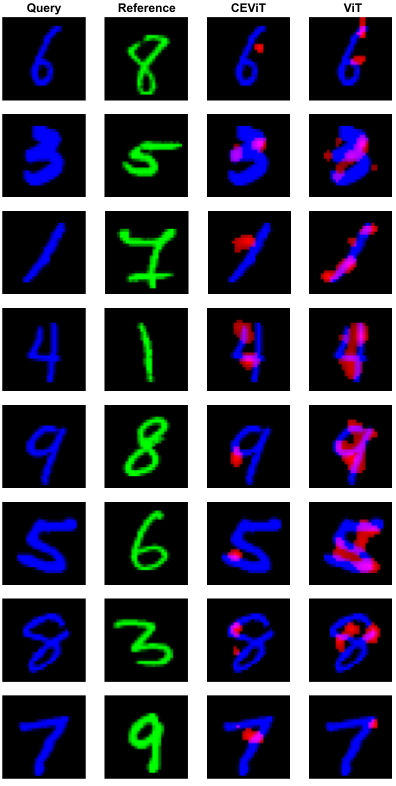}
    \caption{Additional qualitative evaluation samples. All attention masks were sharpened with a threshold of 0.75.}
    \label{fig:qualitative}
\end{figure*}

Figure \ref{fig:qualitative} provides additional qualitative evaluation samples. We conducted sharpening on all attention masks with a threshold of 0.75, similar to Section \ref{section 4.3}. 

\end{document}